\documentclass[10pt,twocolumn,letterpaper]{article}

\usepackage{iccv}
\usepackage{times}
\usepackage{epsfig}
\usepackage{graphicx}
\usepackage{amsmath}
\usepackage{amssymb}
\usepackage{multirow}

\usepackage{caption}

\usepackage{subcaption}
\usepackage{makecell}
\usepackage[pagebackref=true,breaklinks=true,letterpaper=true,colorlinks,bookmarks=false]{hyperref}

\newcommand{\Sai}[1]{\textcolor{red}{\textbf{Sai: #1}}}

\newcommand{\boldstart}[1]{\noindent\textbf{#1}}
\newcommand{\boldstartspace}[1]{\vspace{0.1in}\noindent\textbf{#1}}

\newcommand{\dO}{{o}}
\newcommand{\dP}{{p}}
\newcommand{\dR}{{r}}

\newcommand{\OP}{{o \rightarrow p}}
\newcommand{\PR}{{p \rightarrow r}}
\newcommand{\RP}{{r \rightarrow p}}

\newcommand{\predI}{\bar{I}}
\newcommand{\predA}{\bar{A}}
\newcommand{\predS}{\bar{S}}

\newcommand{\comment}[1]{}

\iccvfinalcopy 

\def\iccvPaperID{4213} 

\ificcvfinal\fi
\begin{document}

\title{Deep CG2Real: Synthetic-to-Real Translation via Image Disentanglement}
\author{
\hspace{0.5cm}
Sai Bi\textsuperscript{1} \hspace{1.4cm}
Kalyan Sunkavalli\textsuperscript{2} \hspace{1.2cm}
Federico Perazzi\textsuperscript{2} 
\\
Eli Shechtman\textsuperscript{2} \hspace{1.2cm}
Vladimir G. Kim\textsuperscript{2} \hspace{1.2cm}
Ravi Ramamoorthi\textsuperscript{1} 
\\
\textsuperscript{1}UC San Diego\hspace{1cm}
\textsuperscript{2}Adobe Research
}

\twocolumn[{
\renewcommand\twocolumn[1][]{#1}
\maketitle
}]

\maketitle
\thispagestyle{empty}

%
\begin{abstract}
We present a method to improve the visual realism of low-quality, synthetic images, e.g. OpenGL renderings. Training an unpaired synthetic-to-real translation network in image space is severely under-constrained and produces visible artifacts. Instead, we propose a semi-supervised approach that operates on the disentangled shading and albedo layers of the image. Our two-stage pipeline first learns to predict accurate shading in a supervised fashion using physically-based renderings as targets, and further increases the realism of the textures and shading with an improved CycleGAN network. Extensive evaluations on the SUNCG indoor scene dataset demonstrate that our approach yields more realistic images
compared to other state-of-the-art approaches. Furthermore, networks trained on our generated ``real'' images predict more accurate depth and normals than domain adaptation approaches, suggesting that improving the visual realism of the images can be more effective than imposing task-specific losses.
\end{abstract}

\vspace{-0.5cm}
\section{Introduction}


\begin{figure}[t]
  \centering
  \includegraphics[width=0.98\linewidth]{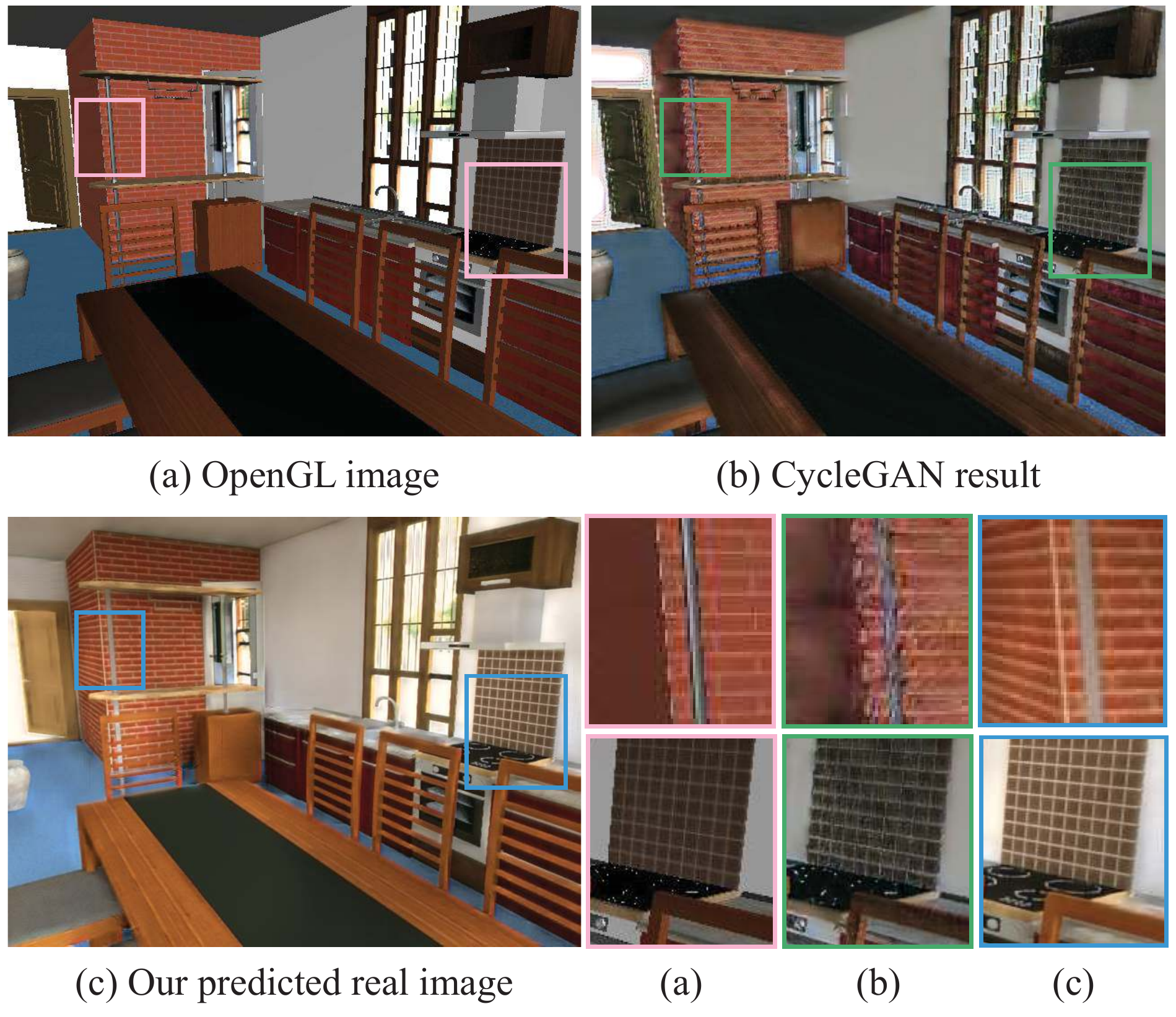}
  \vspace{-0.3cm}
  \caption{Our two-stage adversarial framework translates an OpenGL rendering (a) to 
     a realistic image (c). Compared to single-stage prediction
      with CycleGAN (b), our result has more realistic illumination and better preserves texture details, as shown in the insets. (Best viewed in digital).}
        \vspace{-0.6cm}
\label{fig:teaser}
\end{figure}

Deep learning-based image synthesis methods are generating images with increasingly higher visual quality~\cite{goodfellow2014gan,isola2017pix2pix,karras2018progressive,wang2018pix2pixhd, wang2016generative} even from minimal input like latent codes or semantic segmentation maps. While impressive, one challenge with these approaches is the lack of fine-grained control over the layout and appearance of the synthesized images. On the other hand, it is possible to compose 3D scenes with a desired layout and appearance and render them to create photo-realistic images. However, this requires high-quality scene assets (geometries, materials, lighting) and compute-heavy physically-based rendering.

The goal of this work is to combine the advantages of these two approaches. Given a low-quality synthetic image of a scene---the coarse models in the SUNCG indoor scene dataset~\cite{song2017suncg} rendered with a simple rendering engine like OpenGL---we train a deep neural network to translate it to a high-quality realistic image. One approach to this problem would be to train an unpaired image-to-image translation network, like CycleGAN~\cite{zhu2017cyclegan}, from synthetic OpenGL images to real photographs. However, the unpaired nature of this problem---it is not clear how we would create a dataset of synthetic images with their ``real" counterparts---makes it challenging and results in images with signficiant artifacts like implausible lighting and texture distortions (Fig.~\ref{fig:teaser}(b)). In contrast, our result retains the layout and coarse appearance of the original image, but introduces realistic lighting including global illumination, improves the quality of the scene textures, and removes effects like aliasing (Fig.~\ref{fig:teaser}(c)). 

Improving the realism of a synthetic image requires improving the quality of both illumination and texture. Moreover, these two aspects need to be handled in different ways: illumination needs to be synthesized globally while textures can be modified locally. To this end, we propose disentangling a synthetic image into its constituent shading and albedo layers---i.e., an intrinsic image decomposition~\cite{barrow1978intrinsic}---and training \emph{separate} translation networks for each of them.

We leverage the intrinsic images disentanglement to change this problem from a purely unpaired setting to a two-stage paired-unpaired setting. We render the synthetic scenes with a physically-based renderer (``PBR") to simulate realistic illumination and create paired OpenGL-PBR shading data. In the first stage of our pipeline, we use this data to train an OpenGL-to-PBR shading translation network that synthesizes realistic shading. We combine this new shading with the original OpenGL textures to reconstruct our intermediate PBR images. 

In the second stage, we translate these PBR images to the real image domain in an unsupervised manner using a CycleGAN-like network. We train individual PBR-to-real generators for the shading and albedo layers; we use an encoder-decoder architecture for shading to increase global context, and a purely convolutional network, with no downsampling/unsampling for albedo. As in CycleGAN, the quality of this translation is best with a backward real-to-PBR cycle, and a cycle-consistency loss. In the absence of the disentangled shading-albedo layers for real images, we accomplish this with an asymmetric architecture and a PBR-domain intrinsic image decomposition network.

While our focus is on improving visual quality, our method can be used for domain adaptation. Deep networks can be trained on large-scale, labeled synthetic datasets~\cite{richter2016gta,zhang2017pbr} and prior work has looked at adapting them to improve their performance on real data~\cite{gecer2018face, shrivastava2017learning, mueller2018hands}. Many of these methods impose a task-specific loss on this adaptation~\cite{hoffman2018cycada,murez2018domain,zheng2018t2net}. In contrast, we show that by improving the overall visual realism of synthetic data, we can achieve similar improvements in real image performance on tasks such as normal and depth estimation without the need for task-specific losses, as demonstrated in Table~\ref{table:normal} and Table~\ref{table:depth}.

\begin{figure*}[t]
    \includegraphics[width=\textwidth]{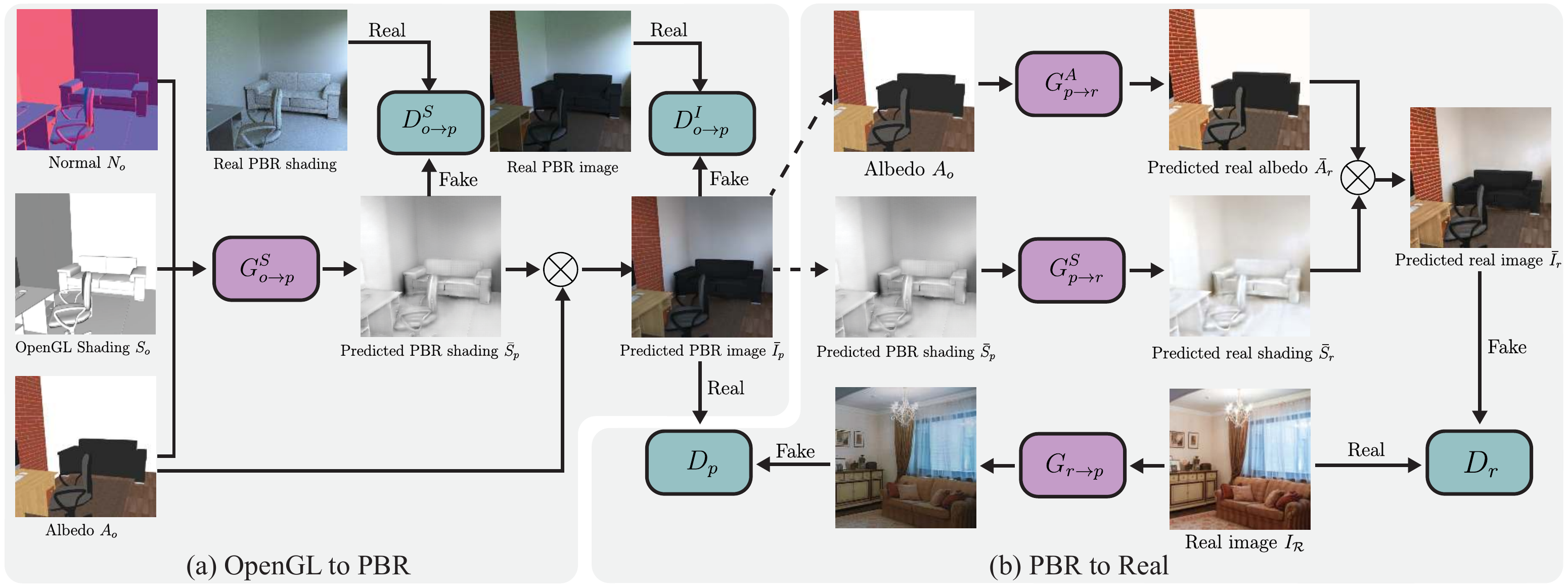}
    \vspace{-0.6cm}
    \caption{The framework of our two-stage OpenGL to real translations.}
    \label{fig:framework}
    \vspace{-0.6cm}
\end{figure*}

\section{Related Work}

\boldstart{Image-to-Image Translation.} 
To improve the realism of synthetic images, Johnson et al.~\cite{johnson2010cg2real} 
retrieve similar patches to the input from real image collections to synthesize realistic imagery. 
Recently deep neural networks have been widely used for this task.
When paired training data is available, previous methods have proposed training conditional generative models with a combination of supervised reconstruction losses and adversarial losses (pix2pix~\cite{isola2017pix2pix}, StackGAN~\cite{zhang2017stackgan}). Such mappings are challenging to learn in the unsupervised setting with only adversarial losses, and prior work has utilized cycle-consistency losses (CycleGAN~\cite{zhu2017cyclegan}) or a shared latent space for the two domains (UNIT~\cite{liu2017unit}). These approaches have been extended to handle multi-modal output (BicyleGAN~\cite{zhu2017multimodal}, MUNIT~\cite{huang2017multimodal}), multiple domains~\cite{choi2018stargan}, higher-resolution images (pix2pix-HD~\cite{wang2018pix2pixhd}), and videos~\cite{wang2018vid2vid}. These methods---especially the unsupervised approaches---can introduce undesired structural changes and artifacts when there is a large domain difference, as there is between synthetic OpenGL images and real images. We handle this by working in a disentangled shading-albedo space. This allows us to a) use a two-stage pipeline that goes from OpenGL images to PBR images and then to the real domain, and b) design separate shading and albedo networks to avoid artifacts.

\boldstart{Domain adaptation.} Domain adaptation methods seek to generalize the performance of a ``task'' network, trained on one domain, to another domain; for example, to train networks on large-scale labeled synthetic datasets and apply them on real images. Feature-space domain adaptation methods either match the distributions of source and target domain features~\cite{sun2016coral} or learn to produce domain-agnostic features using feature-space adversarial losses~\cite{ganin2015unsupervised,ganin2016domain,tzeng2017adversarial}.

Instead, image-space domain adaptation methods seek to match image distributions. The key challenge here is to avoid changing image content in ways that will impair the performance of the task network. This can be handled by using paired source-target data to regularize the translation~\cite{gecer2018face}. In the unpaired setting, prior work has constrained the translated images to be close to the source images~\cite{shrivastava2017learning} but this only works for small domain shifts. Most current methods use a combination of task-specific losses (i.e., preserving the task network's output after translation)~\cite{mueller2018hands}, image-space and feature-space adversarial losses, cycle-consistency losses, and semantic losses (i.e., preserving the semantics of the image after translation)~\cite{hoffman2018cycada,murez2018domain,zheng2018t2net}. 

Our contributions are orthogonal to this direction of work. We demonstrate that translating images in the shading-albedo space leads to higher visual quality, which improves performance on real data. We do this without using task-specific or semantic losses and are thus not constrained to a specific task. Our work can be combined with other domain adaptation ideas to further improve results.

\section{Method}

Our goal is to improve the visual realism of a low-quality synthetic image. In particular, we focus on translating images of indoor scenes, $I_\dO$, from the domain of OpenGL-rendered images $\mathcal{O}$, to the domain of real photographs, $\mathcal{R}$. This is an unpaired problem and has to be learned without direct supervision. The translation has to handle two aspects: first, simple rendering engines like OpenGL do not model complex, real-world lighting, and second, synthetic scenes usually do not have realistic real-world materials and textures. We propose handling this by explicitly manipulating the synthetic shading and albedo separately. Moreover, we find that directly translating from OpenGL to real images is challenging due to the large domain gap between them; changing OpenGL shading to real-world shading requires large non-local transformations that are challenging to learn in an unsupervised manner. However, synthetic scenes can be rendered with physically-based renderers that can generate more realistic shading. We leverage this to propose a two-stage translation. First, we translate the OpenGL image $I_\dO$ to the physically-based rendering (PBR) domain, $\mathcal{P}$, by transforming only the shading using paired OpenGL-PBR images. We then translate these PBR images to the real domain, $\mathcal{R}$, by refining both the albedo and shading using two separate networks; this is done in an unsupervised manner. Figure~\ref{fig:framework} shows an overview of our two-stage framework, and we describe them in following sections.

\subsection{OpenGL-to-PBR Image Translation}
While rendering scenes with OpenGL is computationally fast, the resulting images have low visual realism (e.g., see Figure~\ref{fig:teaser}(a)). One of the reasons is that standalone OpenGL only supports simple lighting models (such as directional or point lights) and does not model complex lighting effects like global illumination. In comparison, a physically based renderer can generate images with photorealistic shading by simulating the propagation of light through a scene, albeit at the cost of significantly more processing time. Therefore, we render the same scene with OpenGL and a physically based renderer Mitsuba~\cite{mitsuba} respectively. Both these images have the same geometry and material properties, and differ only in the quality of the shading. We train a neural network on these image pairs to translate the OpenGL image to the PBR domain; this network thus learns to synthesize more realistic shading from a simple OpenGL scene, bypassing the cost of physically-based rendering.  

We use paired conditional generative adversarial networks~\cite{isola2017pix2pix, wang2018pix2pixhd} for this task. Let $I_\dO, I_\dP$ be a pair of images in the OpenGL and PBR domains, respectively. Since both the images differ only in their shading components, we decompose them into albedo and shading layers and train a shading generator, $G^S_{\OP}(\cdot)$, that transforms OpenGL shading, $S_{\dO}$, to PBR shading, $S_{\dP}$. Instead of using only the OpenGL shading as input, we use the auxiliary buffers of the synthetic scene including albedo $A_\dO$ and surface normals, $N_{\dO}$, as the input to the generator. These additional buffers encode semantic (via albedo) and geometric (via normals and shading) information about the synthetic scene and aid in the translation. Finally, we multiply the synthesized PBR shading, $\predS_\dP$, with the original OpenGL albedo, $A_\dO$, to reconstruct the PBR image
$\predI_{\dP}$:
\begin{equation}
    \predS_{\dP} = G_{\OP}^S(S_\dO, A_\dO, N_\dO),\;\;\; \predI_{\dP} = \predS_\dP * A_\dO
\end{equation}
Our intrinsic image model assumes Lambertian shading. While this is an approximation to real world reflectance, it is sufficient for many regions of indoor images, and is widely used~\cite{bell2014intrinsic,bi2018intrinsic,li2018intrinsic}. To recover high dynamic shading values, we predict the shading in the logarithmic space.

\begin{figure}[t]
  \includegraphics[width=\linewidth]{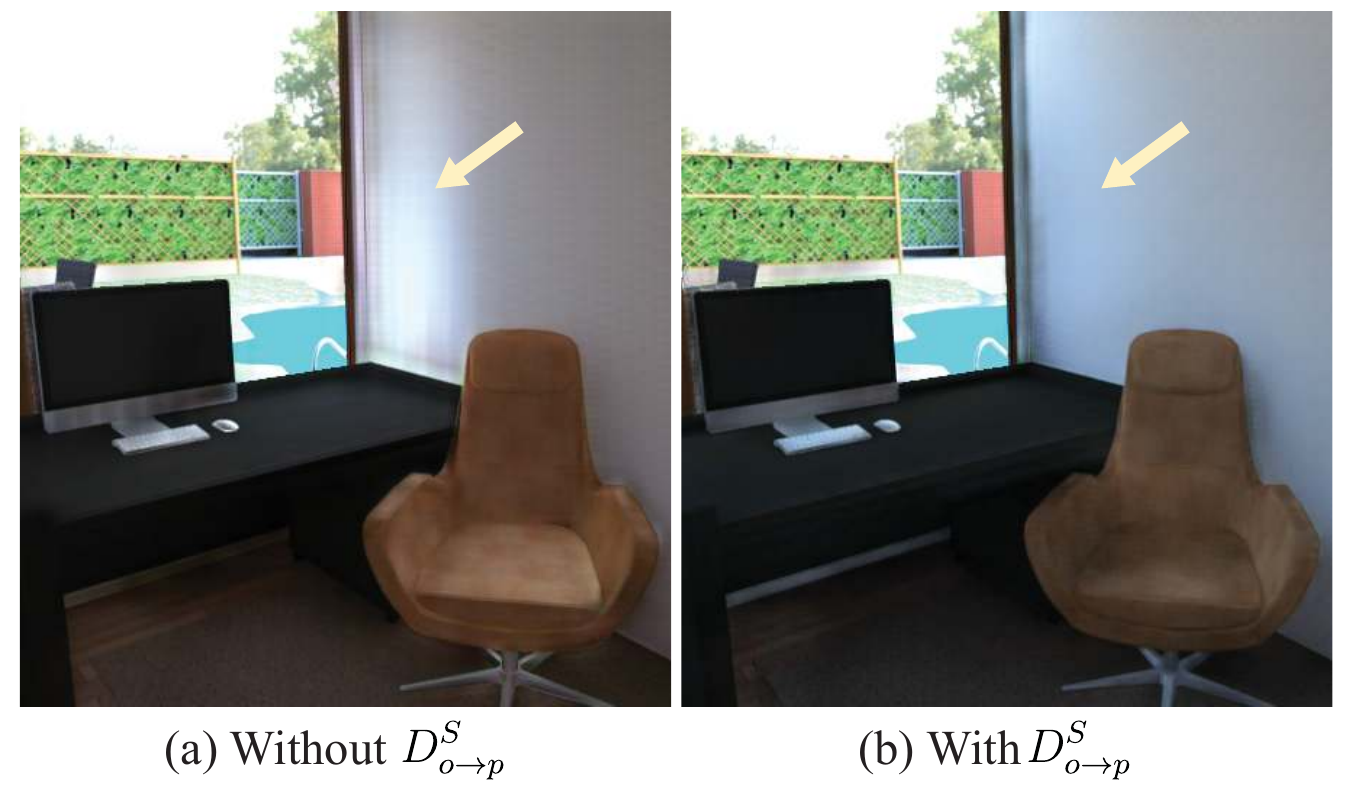}
  \vspace{-0.6cm}
  \caption{With the shading discriminator, our network is able to predict more accurate shading images
  and get rid of problems such as inconsistent colors.}
    \vspace{-0.6cm}
  \label{fig:shading-dis}
\end{figure}

Similar to pix2pixHD~\cite{wang2018pix2pixhd}, we use a combination of a perceptual reconstruction loss (based on VGG features) and adversarial losses. We utilize adversarial losses via a discriminator on the predicted image, $D^I_{\OP}$, as well as one on the predicted shading, $D^S_{\OP}$. This ensures that both the generated images and shadings are aligned with the distribution of PBR images and shading. We used conditional discriminators that also take  albedo and normals as input. 

By translating only the shading and multiplying it with the original albedo, we ensure that the textural structure of the albedo is explicitly preserved. Synthesizing the shading separately is also an easier task for the network because the shading component is spatially smooth and does not have the high-frequency details of the image. Moreover, this also allows us to incorporate the shading discriminator, $D^S_{\OP}$, that provides stronger supervision for training. As shown in Figure~\ref{fig:shading-dis}, without the shading discriminator the network predicts unrealistic illumination with inconsistent colors while our full network avoids this problem.



\subsection{PBR-to-Real Image Translation}

The first stage of our pipeline generates images with more physically accurate shading and global illumination. However, these images may still have a domain shift from real photographs. The albedos are still from the low-quality scenes and may have unrealistic colors. The shading is also still a function of the synthetic scenes and might not align with the intensity, color, contrast and spatial distribution of real-world scenes. In the second stage, we seek to bridge this remaining domain gap to a target real image dataset.

Unlike the first stage, there is no one-to-one correspondence between predicted PBR images and real images. Zhu et al.~\cite{zhu2017cyclegan} introduced the CycleGAN framework to tackle such unpaired problems and we build on their framework for our second stage. Different from the original CycleGAN network which performs the translation in image space, we propose translating the albedo and shading components separately with different generators. This novel design is based on the insight that the albedo contains high frequency details, and should be modified locally to better preserve structural details. On the other hand, shading is a global phenomenon (as it is a function of scene and light source geometry) and should take global context into consideration. Similar observations have been made in~\cite{gershbein1994textures}, which uses different operators for global irradiance and local radiosity.

The output of the first stage is the predicted PBR image, $\predI_\dP$. As noted before, this is the product of the predicted (OpenGL-to-PBR) shading, $\predS_\dP$, and the original scene albedo, $A_\dP$ (which is the same as $A_\dO$). As shown in Figure~\ref{fig:framework}, to translate from the PBR domain to the real domain ($\dP\rightarrow\dR$), we use two
generators $G^A_{\PR}(\cdot)$ and $G^S_{\PR}(\cdot)$ to synthesize the real albedo and shading, $\predA_\dR$ and $\predS_\dR$, respectively. The final predicted real image, $\predI_\dR$, can then be reconstructed as:
\begin{equation}
    \predA_\dR = G^A_{\PR}(A_\dP), \; \predS_\dR = G^S_{\PR}(\predS_\dP), \;\;\; \predI_\dR = \predA_\dR * \predS_\dR
    \label{eqn:stage-2}
\end{equation}

\begin{figure}[t]
        \vspace{-0.3cm}
    \begin{subfigure}{.5\linewidth}
        \centering
        \includegraphics[width=0.98\linewidth]{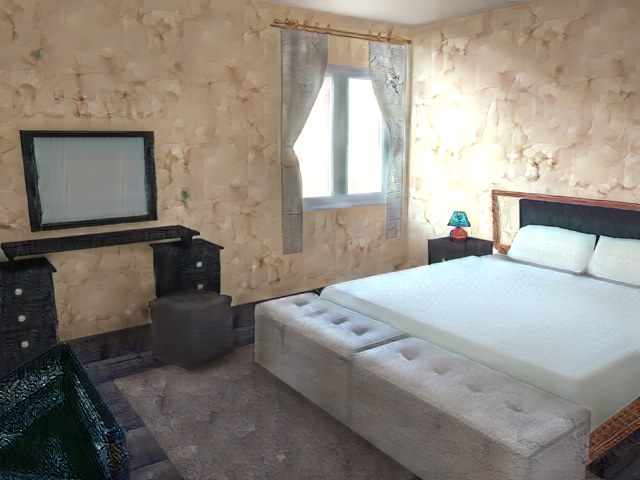}
        \subcaption{With downsampling} 
    \end{subfigure}%
    \begin{subfigure}{.5\linewidth}
        \centering
        \includegraphics[width=0.98\linewidth]{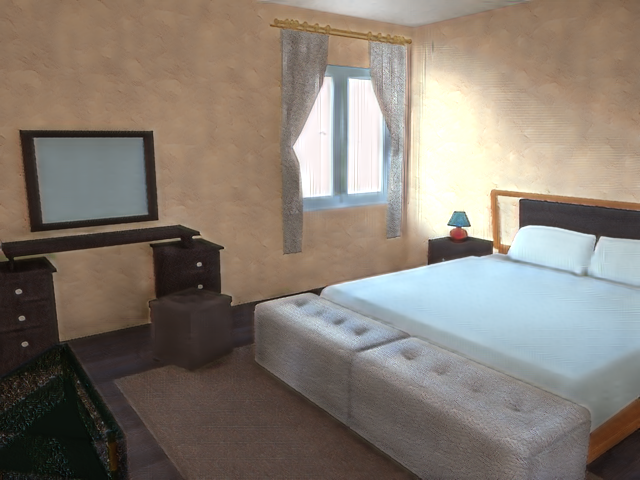}
        \subcaption{Without downsampling}
    \end{subfigure}%
          \vspace{-0.3cm}
    \caption{Comparison between with and without downsampling blocks in the albedo generator. From the result, we can see that 
       the generator without downsampling blocks could better preserve the texture structures of the input.}
       \vspace{-0.3cm}
    \label{fig:albedo-arch-comp}
\end{figure}

We use different architectures for $G^A_{\dP}$ and $G^S_\dP$. For the albedo, we use a fully convolutional network without downsampling or upsampling blocks. This results in a small receptive field for the network and better preserves the texture details while avoiding large structural changes~\cite{ignatov2017dslr, ignatov2018wespe}. As shown in Figure~\ref{fig:albedo-arch-comp}, allowing downsampling blocks in the albedo generator leads to serious high-frequency artifacts in the textures. Our architecture removes this problem and achieves results with higher quality. In contrast, the shading generator uses downsampling blocks for a larger receptive field in order to allow global changes. 

\begin{figure}[t]

    \includegraphics[width=0.98\linewidth]{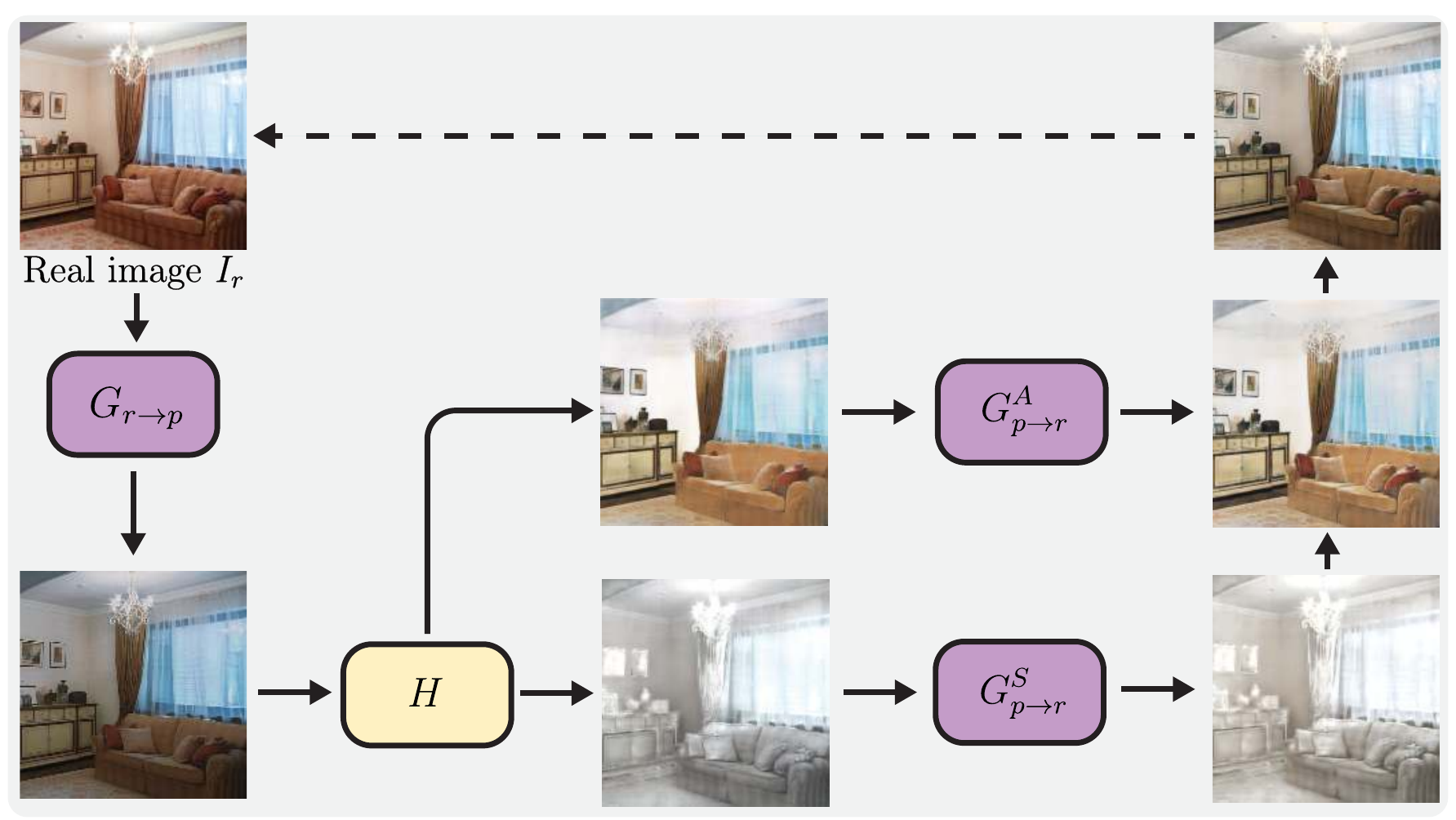}
    \vspace{-0.3cm}
    \caption{To finish the backward cycle, the real image is first translated to the PBR domain. 
    Afterwards we use the pretrained intrinsic decomposition network $H$ to decompose it into its albedo and 
shading, which are further fed to corresponding generators. Finally we multiply 
the output albedo and shading to reconstruct the original real image.}
    \label{fig:backward}
    \vspace{-0.5cm}
\end{figure}

Similar to CycleGAN, we constrain the forward $\PR$ translation using a backward $\RP$ translation. Unlike the PBR domain, we do not have access to albedo and shading layers for real images. Therefore, we use an image-space generator $G_\RP(\cdot)$ that transforms real images to the PBR domain. We utilize two discriminators, $D_\dR$ and $D_\dP$, to distinguish real and fake samples in real and PBR domains respectively. Here, $D_\dR$ distinguishes between the PBR images translated to the real domain ($\predI_\dR$ from Equation~\ref{eqn:stage-2}) and real images. $D_\dP$ on the other hand discriminates between PBR image (synthesized from the first stage), $\predI_\dP$, and real images translated to the PBR domain, $G_\RP(I_\dR)$. Note that while the generators for the $\PR$ direction are applied to albedo and shading, the discriminator is applied to the image computed as their product. We train the network by optimizing the standard GAN loss $\mathcal{L}_{\text{GAN}}(G^A_\PR, G^S_\PR, D_\dR)$ and $\mathcal{L}_{\text{GAN}}(G_\RP, D_\dP)$ for the forward translation $\PR$ and backward translation $\RP$, respectively.

Only having the GAN loss is not sufficient for learning meaningful translations because of the lack of pixel-level correspondence~\cite{yi2017dualgan}. Similar to CycleGAN, we also include forward and backward cycle consistency losses. The forward cycle consistency, ${p \rightarrow r \rightarrow p}$ is trivial: we project the predicted real image $\predI_\dR$ back to the PBR domain by feeding it to the generator $G_{\RP}$, and minimize the $L_1$ loss between the output and the PBR source image $\predI_\dP$:
\begin{align}
    \mathcal{L}_{\text{for}}(G^A_\PR, G^S_\PR, G_\RP) = ||G_\RP(\predI_\dR) - \predI_\dP||_1 \nonumber \\
    = ||G_\RP(G^A_\PR(A_p)*G^S_\PR(\predS_p)) - \predI_\dP||_1
    \label{eqn:forward}
\end{align}

Specifying the backward cycle is more challenging. We can map a real image $I_\dR$ to the PBR domain as $G_\RP(I_\dR)$. However, translating it back to the real domain via the forward $\PR$ translation requires an albedo-shading separation that we do not have for these images---we only have them for our original synthetic images. We tackle this by training an intrinsic decomposition network~\cite{li2018intrinsic, bi2018intrinsic, cheng2018intrinsic} to predict the albedo and shading layers for PBR images. 

Let $H$ be the intrinsic decomposition network. Given a PBR image $I$ and corresponding albedo $A_I$ and shading $S_I$, $H$ is trained by optimizing the following loss function:
\begin{align}
    \mathcal{L}(H) = ||H_A(I) - A_I||_2^2 +||H_S(I) - S_I||_2^2, 
\end{align}
where $H_A(I)$ and $H_S(I)$ are the predicted albedo and shading by the network $H$. We adopt the network architecture used in Li et al.~\cite{li2018intrinsic}, which contains one encoder and two decoders with skip connections. While they use a scale-invariant mean square error (MSE) loss, we use MSE because we require the product of the albedo and shading to be identical to the original image. The intrinsic decomposition network $H$ is pretrained on the predicted PBR images from the first stage, $\predI_\dP$, where we have ground truth albedo and shading. Afterwards, it is fixed during the training of the image translation networks. 

We use the intrinsic decomposition network, $H$, to decompose $G_{\RP}(I_\dR)$ into its albedo and shading. Then we can translate each component via the forward $\PR$ translation to synthesize the result of the full backward cycle, leading to the following backward cycle consistency loss: 
\begin{align}
    &I_\dP' = G_\RP(I_\dR) \nonumber \\
    &I_\dR'' = G^S_\PR\big(H_S(I_\dP')) * G^A_\PR(H_A(I_\dP'))\big) \nonumber \\
    &\mathcal{L}_{\text{back}}(G^A_\PR, G^S_\PR, G_{\RP}) = || I_\dR'' - I_\dR||_1
    \label{eqn:backward}
\end{align}
Figure~\ref{fig:backward} shows the formulation of our backward cycle.

\begin{figure}[t]
    \begin{subfigure}{.5\linewidth}
        \centering
        \includegraphics[width=0.98\linewidth]{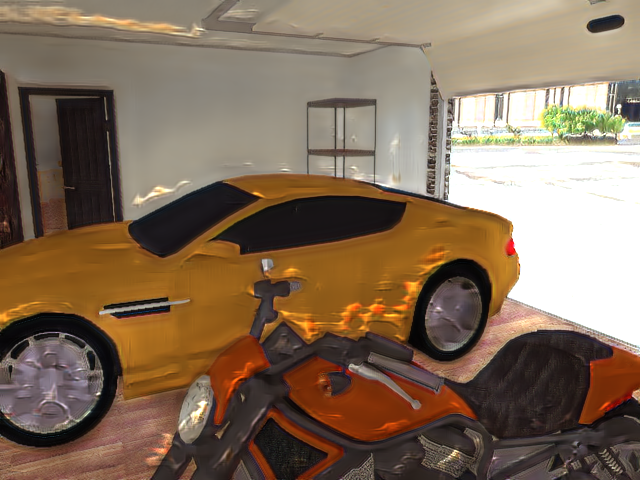}
        \subcaption{Without $\mathcal{L}_{\text{back}}$}
    \end{subfigure}%
    \begin{subfigure}{.5\linewidth}
        \centering
        \includegraphics[width=0.98\linewidth]{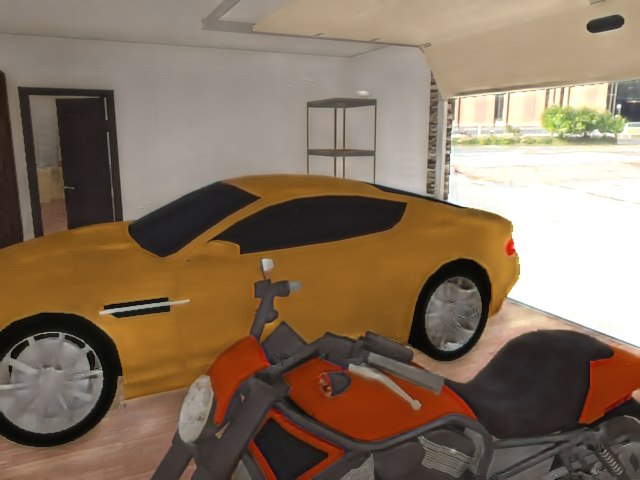}
        \subcaption{With $\mathcal{L}_{\text{back}}$}
    \end{subfigure}%
    \vspace{-0.3cm}
    \caption{Without the backward cycle, the network tends to generate outputs with undesired new structures.
    Adding the backward cycle with a pretrained intrinsic decomposition network is able to generate images with higher quality.}
    \label{fig:comp_back_cycle}
\end{figure}

Note that our network is asymmetric; the forward translation takes our PBR albedo and shading layers, translates and combines them to construct the result. Our backward translation does the opposite; it first translates real images to the PBR domain and then decomposes them there. This allows us to bypass the requirement for real albedo-shading data, and instead train a PBR albedo-shading decomposition \emph{for which we have ground truth supervision}. As can be seen in Figure~\ref{fig:comp_back_cycle}, utilizing this novel backward consistency cycle significantly reduces artifacts and improves the visual quality of our PBR-to-real translations.

\comment
{
In summary, the final loss function for the PBR-to-real translation is as follows:
\begin{align}
    \mathcal{L}(&G^A_\dP, G^S_\dP, G_{\dR}, D_{\dP}, D_{\dR}) = 
  \mathcal{L}_{\text{GAN}}(G^A_\dP,G^S_\dP, D_{\dR}) \nonumber \\
  & + \mathcal{L}_{\text{GAN}}(G_{\dR}, D_{\dP})
  + \lambda_{\text{cyc}} \mathcal{L}_{\text{forward}}(G^A_\dP,G^S_\dP,
  G_{\dR}) \nonumber \\
  & + \lambda_{\text{cyc}} \mathcal{L}_{\text{backward}}(G^A_\dP, G^S_\dP, G_{\dR}), 
\end{align}
where $\lambda_{cyc}$ is the weight for the cycle consistency loss.
 }
Our final loss function for PBR-to-real translation is a combination of the two GAN losses (from discriminators $D_{\dP}$ and $D_{\dR}$), and the forward and backward cycle consistency losses (Equations~\ref{eqn:forward} and \ref{eqn:backward}). 

\section{Implementation}\label{sec:implementation}
\boldstart{OpenGL-to-PBR.} Our first stage network architecture is based on pix2pixHD~\cite{wang2018pix2pixhd}. We use a $70 \times 70$ PatchGAN~\cite{isola2017pix2pix} for the discriminator.  The generator $G^S_\OP$ contains two sub-nets including a global and a local network each with $9$ and $3$ residual blocks as proposed by Wang et al.~\cite{he2015deep}. 
While the shading directly predicted by the generator is reasonable, it may have noise, discontinuities or typical GAN blocking artifacts, which degrade the finale image quality. We leverage the inherent spatial smoothness of the shading and remove potential artifacts by applying a \textit{guided filter layer}~\cite{wu2018fast} after the output layer of the shading generator. The guided filter layer $h(\cdot)$ takes the OpenGL shading, $S_\dO$, and translated OpenGL shading, $G^S_{\OP}(S_\dO)$, as input, and outputs the predicted shading, $\predS_\dP$. We set radius to $4$ and regularization parameter to $0.01$.

\boldstart{PBR-to-Real.} For our second stage, we use the same PatchGAN architecture for the discriminator. Both the shading generator $G^S_\PR$ and real-to-PBR generator $G_\RP$ have $2$ convolutional blocks with stride of $2$ to downsample the input, followed by $4$ residual blocks and $2$ convolutional blocks with upsampling layers. The downsampling/upsampling gives the shading generator a large receptive field allowing it to capture global information about the scene and make global changes such as adjusting shading colors and intensity. The albedo generator $G^A_\PR$ has a similar architecture as $G^S_\PR$, except that it does not have any downsampling/upsampling. This keeps the receptive field small and forces the albedo generator to only modify the albedo locally thereby preserving textural details.

\boldstart{Training data.} For the OpenGL and PBR images we use the synthetic dataset from Li et al.~\cite{li2018intrinsic}, which contains about $20000$ $480 \times 640$ images of indoor scenes from the SUNCG dataset~\cite{song2017suncg}, rendered with both OpenGL and Mitsuba~\cite{mitsuba}, a physically based renderer.  We use $5000$ image pairs for training the first stage. Then the first-stage network is used to translate another $5000$ OpenGL images to the PBR domain, which are used for second-stage training. Two real image datasets are used in the second stage: the real estate dataset from Poursaeed et al.~\cite{poursaeed2018vision} and the NYUv2 dataset~\cite{silberman2012nyu}. The real estate dataset contains high quality real images of indoor scenes, and we use it for comparison on image quality (Section~\ref{sec:visual}). The NYUv2 dataset is used for domain adaptation experiments (Section~\ref{sec:domain}). Finally an extra $5000$ OpenGL images are used for testing the full pipeline. We select the training and testing OpenGL images from different scenes to avoid overlap.

\boldstart{Training details.}
Both stages are trained separately. We use Adam~\cite{kingma2015adam} with an initial learning rate of $0.0002$ for training both stages. The networks are trained for $100$ epochs, with the first $50$ epochs trained with the initial learning rate, and the remaining $50$ epochs with a linearly decaying learning rate. We randomly crop patches from the input images for training with a patch size of $400 \times 400$ for the first stage and $256 \times 256$ for the second stage.

\section{Results}
In this section, we first show the comparison on visual quality of translations against 
baseline methods (Section~\ref{sec:visual}).  Afterwards we show that our two stage pipeline could 
boost the performance of network models on real images when trained with our translated images (Section~\ref{sec:domain}). 

\begin{figure}[t]
    \centering
    \begin{subfigure}{.495\linewidth}
        \centering
        \includegraphics[width=0.98\linewidth]{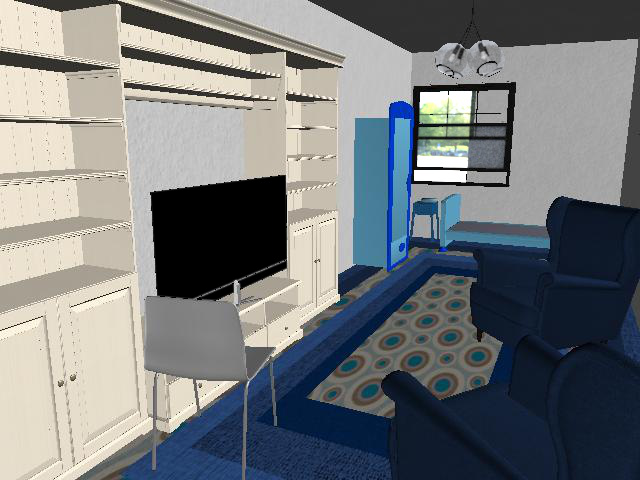}
        \subcaption{OpenGL image}
    \end{subfigure}%
    \begin{subfigure}{.495\linewidth}
        \centering
        \includegraphics[width=0.98\linewidth]{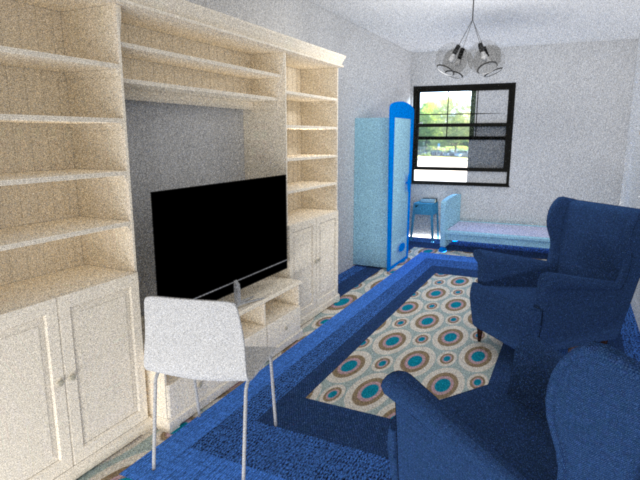}
        \subcaption{PBR image}
    \end{subfigure}%

    \begin{subfigure}{.33\linewidth}
        \centering
        \includegraphics[width=0.98\linewidth]{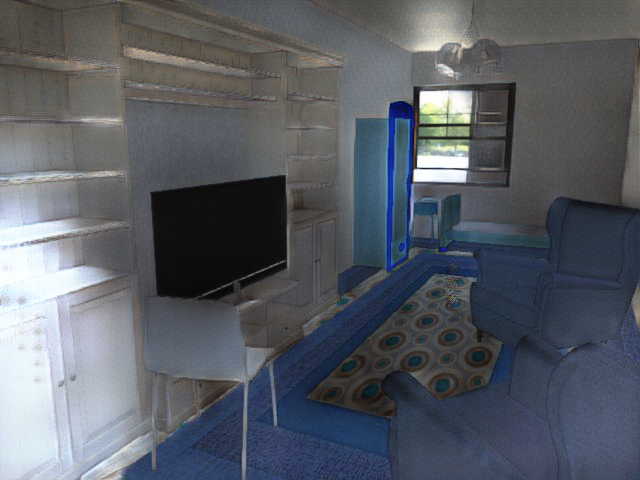}
        \subcaption{p2pHD-OpenGL}
    \end{subfigure}%
    \begin{subfigure}{.33\linewidth}
        \centering
        \includegraphics[width=0.98\linewidth]{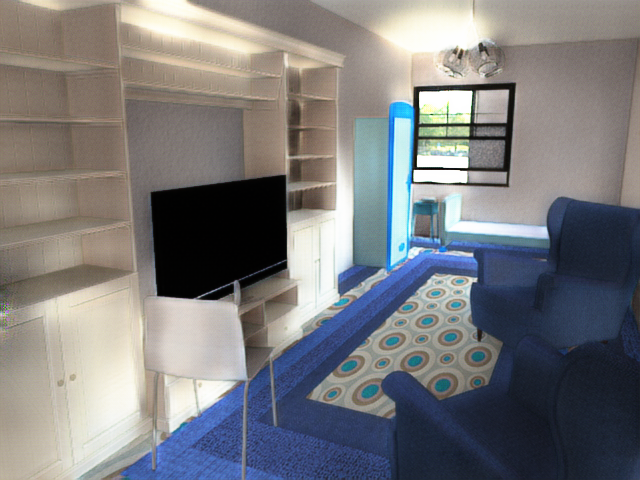}
        \subcaption{p2pHD-S+A+N}
    \end{subfigure}%
    \begin{subfigure}{.33\linewidth}
        \centering
        \includegraphics[width=0.98\linewidth]{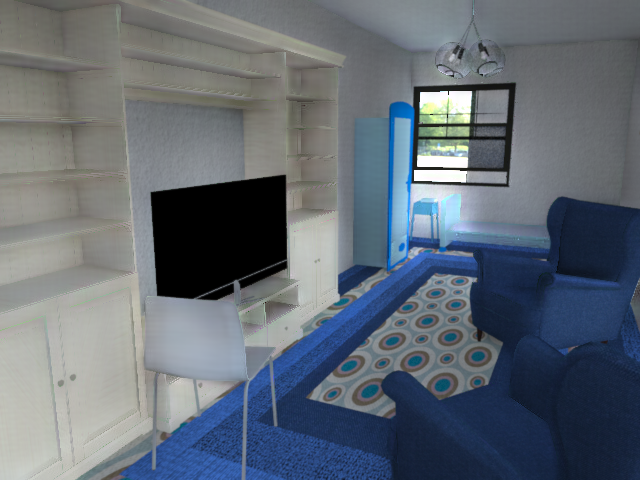}
        \subcaption{Our result}
    \end{subfigure}%

    \vspace{-0.3cm}
    \caption{OpenGL to PBR comparison.
    A PBR renderer like Mitsuba needs around 20 mins. to render a noise-free image as in (b). In comparison our network can generate  a high quality result in $0.03$ secs. (e). 
    Compared to the pix2pix-HD 
    framework that directly 
    predicts the output images using OpenGL (c) or auxiliary buffers (d), with inconsistent shading such as abrupt highlights on the cabinet, our method generates images
    with much higher visual realism. 
    }
       \vspace{-0.3cm}
  \label{fig:first-stage}
\end{figure}

\subsection{Comparison on visual quality}\label{sec:visual}
We compare the results of our method against different baselines for each separate stage and the whole pipeline.
In addition to qualitative comparisons, we also quantitatively measure the visual realism of images 
generated with different methods with two metrics. The first is the FID score~\cite{heusel2017gans}
which has been shown to be effective in measuring 
the distance between two distributions and is consistent with image noise and distortion. 
In addition, we also conduct a human perceptual study on Amazon Mechanical Turk. For each task, 
workers are shown the images output by different methods, and asked to select the most realistic result with the fewest artifacts. For each task, we have 
$100$ images, each evaluated by $10$ different workers, for a total of $1000$ judgements. We discuss each stage in detail below.


\boldstart{OpenGL to PBR translation.} Our proposed network takes the albedo, normal and 
OpenGL shading as input to  predict the PBR shading, which is multiplied by the albedo 
to reconstruct the desired PBR image. We compare our method with two baselines:
the pix2pix-HD network~\cite{wang2018pix2pixhd} that predicts  
the PBR image conditioned on the OpenGL image only (denoted as \textit{p2pHD-OpenGL} in Table~\ref{table:score} ), and the same 
network that takes the same input as ours 
but directly predicts the PBR images (\textit{p2pHD-S+A+N}). Both baselines are trained with the same generator architecture and 
parameter settings as ours. The only differences are the generator inputs/outputs and the lack of shading discriminators. 

We calculate the FID scores between the predicted PBR images and ground truth PBR images, and we also
conduct a user study to ask workers to select the most visually realistic PBR predictions among outputs of 
three methods. From the results in Table~\ref{table:score} ($\mathcal{O}\rightarrow \mathcal{P}$), we can clearly see that our method achieves a much lower 
FID score compared to other baselines, implying that our predictions are more aligned with the distribution
of the ground truth PBR images. In addition, we obtain a user preference rate of $60.6\%$,
much higher than the other two baselines, demonstrating that doing predictions in shading 
space generates images with much higher quality. 
A visual comparison is shown in Figure~\ref{fig:first-stage}.
\comment{Figure~\ref{fig:first-stage} shows a visual comparison 
between our approach and other baselines. From the figure we can see that our result has more realistic
shading and fewer artifacts. In comparison, directly predicting PBR images with either 
OpenGL or auxiliary buffers as input produces inconsistent shading across different regions.}

\begin{table}[t]
\centering
\renewcommand{\arraystretch}{1.0}
\begin{tabular}{c|c|rc}
\hline
\multicolumn{2}{l|}{} &  \makecell{FID } & \makecell{User \\Preference} \\ \hline
\multirow{3}{*}{$\mathcal{O \rightarrow P}$}
    & p2pHD-OpenGL~\cite{wang2018pix2pixhd}  & 21.01 &  10.2\%\\ 
    & p2pHD-S+A+N~\cite{wang2018pix2pixhd} &  11.63 &  29.2\%\\ 
    & \textbf{Ours} &  \textbf{7.40} &  \textbf{60.6\%}\\ \hline
\multirow{2}{*}{$\mathcal{P \rightarrow R}$}
& CycleGAN~\cite{zhu2017cyclegan} &  54.80 & 27.9 \% \\ 
    & \textbf{Ours} & \textbf{53.48} &  \textbf{72.1\%} \\ \hline 
\multirow{3}{*}{$\mathcal{O \rightarrow R}$}
    & CycleGAN~\cite{zhu2017cyclegan} & 59.42  &  16.9\% \\ 
    & T\textsuperscript{2}Net~\cite{zheng2018t2net}&  65.33 & 4.7\%  \\ 
    & \textbf{Ours}& \textbf{53.48}  &  \textbf{78.4\%}\\ \hline
\end{tabular}
\vspace{-0.3cm}
\caption{Comparison on FID and user preference score.}
\label{table:score}
\end{table}

\begin{figure}[t]
 \vspace{-0.4cm}
  \includegraphics[width=\linewidth]{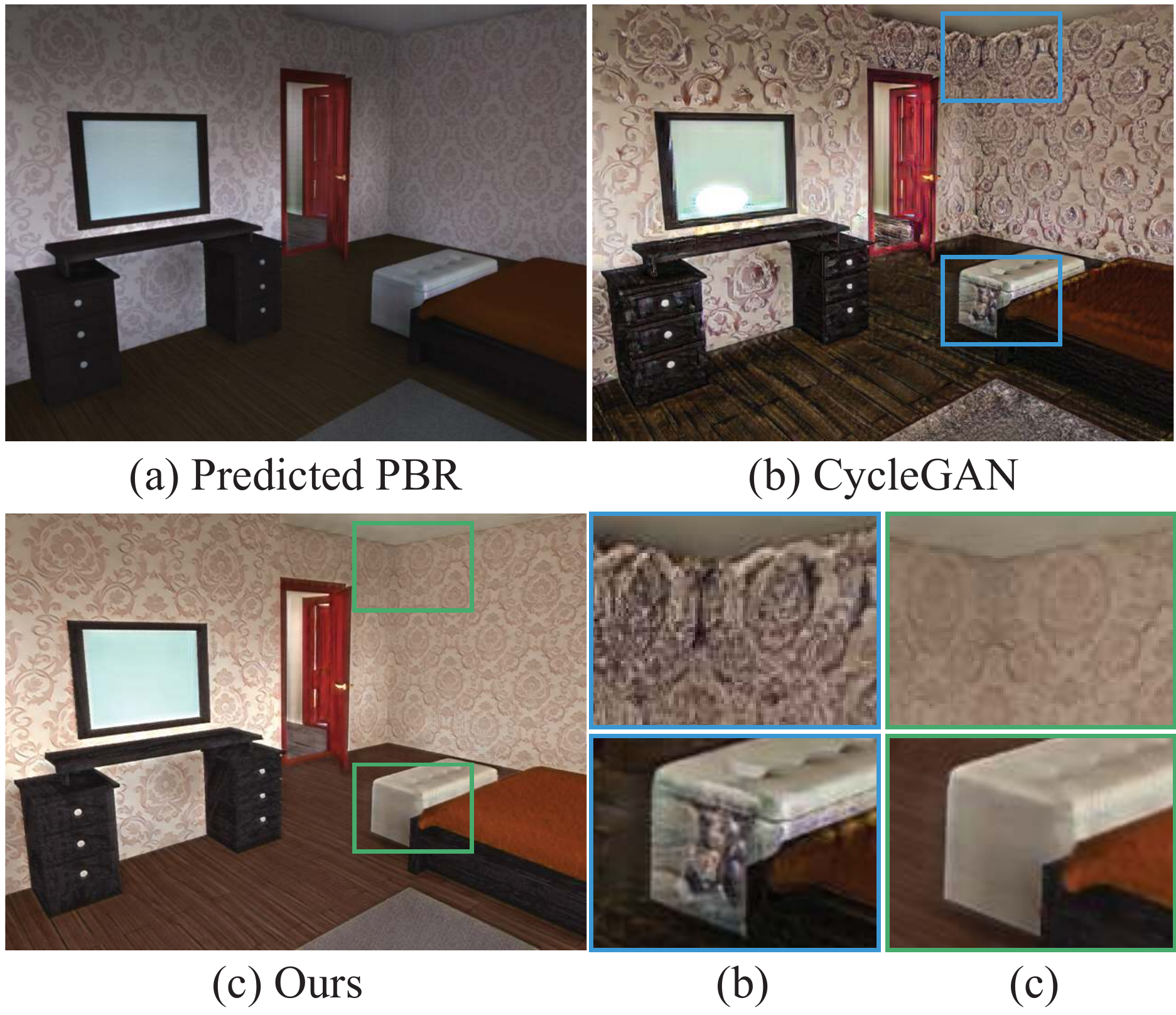}
  \vspace{-0.6cm}
  \caption{PBR to Real comparison. Given the predicted PBR image (a) in the first stage, we further 
  translate it to real domain. Compared to the original CycleGAN~\cite{zhu2017cyclegan} 
    that performs the translation in image space, our method could better preserve the 
structure of the textures and generate images with many fewer artifacts.}
  \label{fig:second-stage}
  \vspace{-0.5cm}
\end{figure}

\begin{figure*}[t]
  \centering
  \vspace{-0.5cm}
  \includegraphics[width=\textwidth]{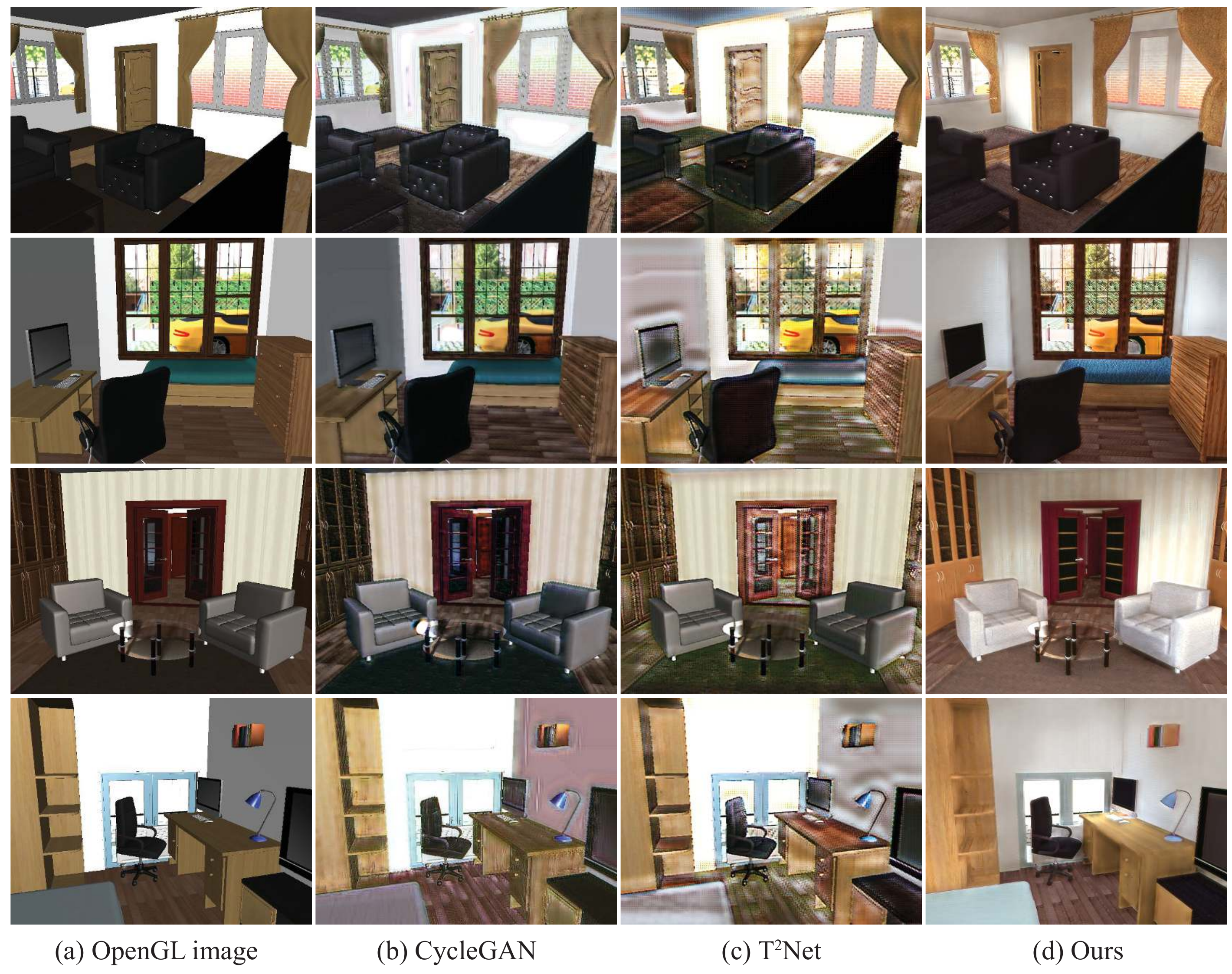}
    \vspace{-0.6cm}
  \caption{Comparison on the full pipeline from OpenGL to Real against single stage translation 
      with CycleGAN and T\textsuperscript{2}Net.}
  \label{fig:full-pipeline}
  \vspace{-0.5cm}
\end{figure*}

\boldstart{PBR to Real translation.} In this stage, we train the network to translate the output of 
the first stage to the real domain. We compare to the na\"ive CycleGAN framework that performs the translations
in image space. FID score is calculated between 
the translated images and real images from the real estate dataset.  Similarly, from Table~\ref{table:score} we can see that 
disentangled translations on albedo and shading with our proposed framework achieves
lower FID scores and much higher user preference rate. As shown in Figure~\ref{fig:second-stage}, 
our network is able to modify the albedo and shading of the input while preserving the texture structures.
In contrast, the original CycleGAN network introduces obvious artifacts which degrade 
image quality.

\boldstart{Full pipeline.} We also compare our two-stage translations to previous methods 
that translate directly from OpenGL to real: the CycleGAN method and T\textsuperscript{2}Net~\cite{zheng2018t2net}.
T\textsuperscript{2}Net uses a single generator and discriminator for synthetic to real translation. In addition to standard GAN loss,
it introduces an identity loss to guarantee that when real images are fed to synthetic generator, the outputs are 
similar to the inputs. The results show that our full pipeline significantly outperform the single-stage 
translations used in both baselines. The visual comparison in Figure~\ref{fig:full-pipeline} 
shows that single-stage translation is not able to produce realistic shading and generates results with 
noise and artifacts due to the large gap between OpenGL images and real images, which our proposed method addresses satisfactorily.

\begin{table}[t]

\renewcommand{\arraystretch}{1.25} 
\resizebox{\linewidth}{!}{%
\begin{tabular}{l|c|ccccc}
\hline
\multicolumn{2}{l|}{} & \multicolumn{2}{c|}{Lower is better} &  \multicolumn{3}{c}{Higher is better(\%)}  \\ \cline{3-7}
\multicolumn{2}{l|}{} & Mean & \multicolumn{1}{l|}{Median}  &  $ < 11.25$ &  $ < 22.5$ & $ < 30$  \\ \hline
\multirow{2}{*}{$\mathcal{O \rightarrow P}$}
       & OpenGL  &  37.73 &  31.91 & 17.93 & 38.36 & 48.52 \\ 
       & \textbf{Ours-PBR} &  34.82 & 27.66  & 21.16  &  42.82 & 53.25 \\ \hline
\multirow{1}{*}{$\mathcal{P \rightarrow R}$}  
       & CycleGAN~\cite{zhu2017cyclegan}&  33.90 &  27.24 &  22.28 & 43.99 & 54.38 \\ \hline
\multirow{3}{*}{$\mathcal{O \rightarrow R}$}
       & CycleGAN~\cite{zhu2017cyclegan}&  36.33 & 30.25 & 19.15  & 39.19  & 50.15 \\ 
       & T\textsuperscript{2}Net~\cite{zheng2018t2net} &  36.93 & 30.93 & 18.93 & 39.23 & 50.49 \\ 
       & \textbf{Ours-full} & \textbf{33.15} & \textbf{26.27} & \textbf{23.52} & \textbf{44.95} & \textbf{55.28} \\ 
       & Real\textsuperscript{*} & 28.18 & 21.95 & 28.14 &52.21 & 62.35\\ \hline
\end{tabular}
}
\vspace{-0.3cm}
\caption{Normal estimation results on NYUv2 dataset.}
\label{table:normal}
\end{table}

\begin{table}[t]

\renewcommand{\arraystretch}{1.25} 
\resizebox{\linewidth}{!}{%
\begin{tabular}{l|c|ccccc}
\hline
\multicolumn{2}{l|}{} & \multicolumn{2}{c|}{Lower is better} &  \multicolumn{3}{c}{Higher is better (\%)}  \\ \cline{3-7}
\multicolumn{2}{l|}{} & RMSE & \multicolumn{1}{l|}{RMSE-log}  &  $ < 1.25$ &  $ < 1.25^2$ & $ < 1.25^3$  \\ \hline
\multirow{2}{*}{$\mathcal{O \rightarrow P}$}
       & OpenGL  &  1.0770 &  0.3873 & 43.53 & 73.60 & 89.75 \\ 
       & \textbf{Ours-PBR} &  1.0293 & 0.3514  & 46.23  &  75.41 & 91.35 \\ \hline
\multirow{1}{*}{$\mathcal{P \rightarrow R}$}  
       & CycleGAN~\cite{zhu2017cyclegan}&  0.9824 &  0.3394 &  48.24 &  78.61& 92.25 \\ \hline
\multirow{3}{*}{$\mathcal{O \rightarrow R}$}
       & CycleGAN~\cite{zhu2017cyclegan}&  1.0328 & 0.3726 & 45.46 & 75.49 & 91.13 \\ 
       & T\textsuperscript{2}Net~\cite{zheng2018t2net} &  1.0085 & 0.3548 & 47.49 & 77.57 &  91.94\\ 
       & \textbf{Our-full} &  \textbf{0.9774} & \textbf{0.3328} &  \textbf{49.59} & \textbf{79.54} & \textbf{93.14} \\  
       & Real\textsuperscript{*} & 0.7070 &  0.2516 & 67.76 & 88.75 &  96.35 \\ \hline
\end{tabular}
}
\vspace{-0.3cm}
\caption{Depth estimation results on NYUv2 dataset. \textit{Real\textsuperscript{*}} is the model trained on 5000 real images from NYUv2.}
\label{table:depth}
\vspace{-0.5cm}
\end{table}

\subsection{Domain adaptation}\label{sec:domain}
Network models trained with synthetic images often performance poorly on real images due to the domain gap.
Our two-stage network improves the visual realism of synthetic images, thereby boosting the performance
of network models trained with the translated images. 
We compare the performance of network models trained with images generated with different methods on two tasks including normal estimation and 
depth estimation. 
We train the task networks on $5000$ images from SUNCG dataset after translation to the real domain, and evaluate them on the test dataset 
in NYUv2, which contains $1449$ images with ground truth normal and depth. 

\boldstart{Normal estimation.}
We use a U-Net~\cite{ronneberger2015u} with $7$ downsampling blocks for the normal estimation task, and apply
the inverse dot product between ground truth normal and predicted normal as the loss function. The network 
is trained with an Adam optimizer for 200 epochs with a learning rate of $2\times10^{-4}$ for the first 100 epochs
and a linearly decaying rate for the remaining 100 epochs. We test the network on the NYUv2 dataset.
In Table~\ref{table:normal}, we report the mean and median angle between predicted normal and ground truth normal, 
as well as the percentage of pixels where the angle is below a certain threshold. From the table we can see compared to 
the model trained on OpenGL images, training on our predicted PBR images significantly reduces the average angle, and translation to the real domain further reduces it to $33.15$, which demonstrates 
that both stages help reduce the domain gap and help improve network performance.  
In addition, compared to the na\"ive CycleGAN in image space for the second stage, our disentangled design is also able to achieve 
higher accuracy. Finally, compared to going directly from OpenGL to real images 
with methods such as CycleGAN and T\textsuperscript{2}Net, our two-stage translation significantly outperforms
them on all metrics. 

\vspace{-0.2cm}
\boldstartspace{Depth estimation.}
We use the network architecture in Zheng et al.~\cite{zheng2018t2net} for depth estimation, and 
adopt the same training protocol. We evaluate the performance of different approaches on the 
NYUv2 dataset using metrics such as relative mean squared error (RMSE) between the prediction and  ground truth, RMSE in log space as well as the percentage of 
pixels whose ratios to ground truth are below a threshold~\cite{eigen2014depth}. Table~\ref{table:depth}
summarizes the scores of different network models. From the table we can see that training 
with our predicted PBR images achieves much higher accuracy than training 
on the synthetic OpenGL images, and our full pipeline further improves the accuracy by translating
the PBR images to real domain. Our two-stage translation is also able to outperform single-stage translation 
with T\textsuperscript{2}Net and CycleGAN and leads to a lower error. 


\vspace{-0.2cm}
\section{Conclusion}
\vspace{-0.2cm}
We propose a novel two-stage framework to translate synthetic OpenGL images to the real domain. We achieve this by manipulating the albedo and shading layers of an image: we first translate them to the PBR domain by training on paired data,
followed by an unsupervised translation to the real domain. We have demonstrated that our approach leads to translations with higher visual quality and better performance for domain adaptation on scene inference tasks. We believe that operating on the albedo-shading decomposition is a way to incorporate physical structure into generative models and would like to explore applications of this idea to other tasks like image synthesis, image editing and style transfer.


\boldstart{Acknowledgements}
This work was supported in part by ONR grant N000141712687, Adobe
Research, a Samsung GRO grant, and the UC San Diego Center for Visual
Computing. Part of this work was done while Sai Bi was an intern at Adobe Research.

{\small
\bibliographystyle{ieee_fullname}
\bibliography{egbib}
}

\end{document}